\documentclass[lettersize,journal]{IEEEtran}
\usepackage{amsmath,amsfonts}
\usepackage{algorithmic}
\usepackage{array}
\usepackage[caption=false,font=normalsize,labelfont=sf,textfont=sf]{subfig}
\usepackage{textcomp}
\usepackage{stfloats}
\usepackage{url}
\usepackage{verbatim}
\usepackage{graphicx}
\usepackage{amsmath}
\usepackage{amssymb,stmaryrd}
\usepackage{bm}
\usepackage{gensymb}
\usepackage{pifont}
\usepackage{booktabs}
\usepackage{multirow}
\usepackage{algorithm}

\makeatletter
\renewcommand{\maketag@@@}[1]{\hbox{\m@th\normalsize\normalfont#1}}%
\makeatother

\def\BibTeX{{\rm B\kern-.05em{\sc i\kern-.025em b}\kern-.08em
    T\kern-.1667em\lower.7ex\hbox{E}\kern-.125emX}}
\usepackage{balance}
\begin{document}
\title{Online IMU-odometer Calibration using GNSS Measurements for Autonomous Ground Vehicle Localization}
\author{Baoshan Song, Xiao Xia, Penggao Yan, Yihan Zhong,\\ Weisong Wen,~\IEEEmembership{Member,~IEEE}, and Li-Ta Hsu,~\IEEEmembership{Senior Member,~IEEE}
\thanks{The authors are with The Department of Aeronautical and Aviation Engineering, Hong Kong Polytechnic University, Hong Kong.  (e-mail: lt.hsu@polyu.edu.hk).}
}

\markboth{Journal of \LaTeX\ Class Files,~Vol.~18, No.~9, September~2020}%
{How to Use the IEEEtran \LaTeX \ Templates}

\maketitle

\begin{abstract}
Accurate calibration of intrinsic (odometer scaling factors) and extrinsic parameters (IMU–odometer translation and rotation) is essential for autonomous ground vehicle localization. Existing GNSS-aided approaches often rely on positioning results or raw measurements without ambiguity resolution, and their observability properties remain underexplored. This paper proposes a tightly coupled online calibration method that fuses IMU, odometer, and raw GNSS measurements (pseudo-range, carrier-phase, and Doppler) within an extendable factor graph optimization (FGO) framework, incorporating outlier mitigation and ambiguity resolution. Observability analysis reveals that two horizontal translation and three rotation parameters are observable under general motion, while vertical translation remains unobservable. Simulation and real-world experiments demonstrate superior calibration and localization performance over state-of-the-art loosely coupled methods. Specifically, the IMU-odometer positioning using our calibrated parameters achieves the absolute maximum error of 17.75 m while the one of LC method is 61.51 m, achieving up to 71.14\% improvement. To foster further research, we also release the first open-source dataset that combines IMU, 2D odometer, and raw GNSS measurements from both rover and base stations. To support future work, we release the first open-source dataset combining IMU, 2D odometer, and raw GNSS measurements from rover and base stations.
\end{abstract}


\section{Introduction}
\textit{\textbf{Localization for autonomous ground vehicles:}} Localization is a fundamental requirement for AGV, supporting intelligent transportation applications such as delivery, patrolling, search, and rescue \cite{reid_localization_2019}. An IMU and an odometer are two common sensors to provide acceleration, velocity and angular velocity for navigation \cite{liu_visual-inertial_2019}. Generally, they are less susceptible to environmental changes and can be used as dead-reckoning sensors which can incorporate other external sensors (e.g., camera \cite{zuo_visual-inertial_2019}, light detection and ranging (LiDAR) \cite{zhang_lidar-imu_2019} and GNSS \cite{gao_odometer_2018}) to achieve driftless positioning. The problem is, these external sensors are sensitive to environmental conditions. For examples, the camera is affected by light changes; LiDAR needs signal reflection from blocks; and GNSS is degraded in urban canyons due to multipath effects and none-line-of-sight (NLOS) receptions. Consequently, it is critical to enhance the robustness of the IMU and the odometer fused dead-reckoning method (e.g., IMU-odometer) \cite{chen_estimate_2021}. To meet the need of the dead-reckoning, accurate calibration of ego-sensor intrinsic parameters and inter-sensor extrinsic parameters are prerequisites to reduce the effect of systematic errors. For examples, inaccurate IMU bias, odometer scaling factors and extrinsic spatial transformation between IMU-odometer may cause rapid drift \cite{li_high-precision_2023,georgy_modeling_2010}.  Thus, their calibrations are required.

\textit{\textbf{Camera and LiDAR aided IMU-odometer calibration is feasible but could drift in long term:}} In recent years, visual sensors and LiDAR have became popular in navigation systems and they are also used to calibrate the IMU-odometer parameters \cite{huang_visual-inertial_2019}. Based on the baseline like VINS \cite{qin_robust_2017}, MSCKF \cite{mourikis_multi-state_2007} and LOAM \cite{zhang_loam_2014}, the camera and the LiDAR have been employed to aid IMU-odometer calibration. These methods leverage sliding-window estimation and tightly coupled measurements to mitigate nonlinearity and non-Gaussian noise. Variations between these works exist in the number of encoders and motion models used, ranging from simple 1D single-wheel models \cite{liu_visual-inertial_2019, zhang_lidar-imu_2019, huang_visual-inertial_2019, zhao_online_2022, zhao_vehicle-motion-constraint-based_2023, zhang_mounting_2024}, 2D multi-encoder systems \cite{zuo_visual-inertial_2019, zuo_visual-based_2024, lee_mins_2023, xiong_g-vido_2022} and even barometer-augmented 3D setups \cite{che_visualinertialwheel_2025}. Specifically, a 1D single-wheel type is to employ a single wheel encoder to provide forward linear velocity \cite{zhao_vehicle-motion-constraint-based_2023}. After that, some authors employ various models, such as skid-steering, differential and instantaneous centers of rotation (ICR) \cite{zuo_visual-inertial_2019}, to fuse measurements from more encoders to generate 2D measurements. Recently, barometer is integrated in the system to provide measurement in altitude to provide full 3D measurements with multiple wheel encoders \cite{che_visualinertialwheel_2025}. However, a barometer is usually used on aerial vehicles but not common on ground vehicles.
Observability analysis has also been conducted in some works to understand calibration mechanisms \cite{zuo_visual-inertial_2019,zhao_vehicle-motion-constraint-based_2023,zuo_visual-based_2024}. With the help of a camera, the extrinsic and intrinsic of related to 2D odometer can be calibrated.
Despite these advances, camera and LiDAR aided methods are limited by their properties: (1) visual features are not always reliable in all environments due to light changes and LiDAR measurements could be limited by poor geometry; (2) camera and LiDAR measuring relies on precise intrinsic parameters and is sensitive to camera hardware configuration; and (3) camera and LiDAR aided calibration based on local odometry could drift after long-term operation in large-scale environments. These limitations lead to the first research gap.

\begin{table*}[t]
\centering
\caption{Related works on sensor-aided IMU-odometer calibration}
\label{tab:related_works}
\begin{tabular}{p{2cm} p{2cm} p{1cm} p{1cm} p{1cm} p{1cm} p{1cm} p{1cm}}
\toprule
Representative Work & Supported Sensors & Odometer Meas. & Estimator & Couple Mode & Open-source & Amb. Res. & Obs. Analysis \\
\midrule
Liu \cite{liu_visual-inertial_2019},  Zhao \cite{zhao_online_2022}, Zhang \cite{zhang_mounting_2024} & camera-IMU-odometer    & 1D & Graph  & TC & \ding{55} & – & \ding{55} \\
Zhao \cite{zhao_vehicle-motion-constraint-based_2023} & camera-IMU-odometer    & 1D & Graph  & TC & \ding{55} & – & \ding{51} \\
Zuo \cite{zuo_visual-inertial_2019,zuo_visual-based_2024}    & camera-IMU-odometer    & 2D & Graph  & TC & \ding{55} & – & \ding{51} \\
Lee \cite{lee_mins_2023} & camera-LiDAR-IMU-odometer & 2D & Graph  & TC & \ding{51} & – & \ding{55} \\
Zhang \cite{zhang_lidar-imu_2019}        & LiDAR-IMU-odometer     & 1D & Filter & TC & \ding{55} & – & \ding{55} \\
Sun \cite{sun_uwbimuodometer_2025}          & UWB-IMU-odometer       & 2D & Graph  & TC & \ding{55} & – & \ding{55} \\
Wu \cite{wu_self-calibration_2010}           & GNSS-IMU-odometer      & 1D & Filter & LC & \ding{55} & \ding{55} & \ding{51} \\
Bai \cite{bai_improved_2021}          & GNSS-IMU-odometer      & 2D & Graph  & LC & \ding{55} & \ding{55} & \ding{55} \\
Bai \cite{bai_graph-optimisation-based_2022,bai_enhanced_2023}   & GNSS-IMU-odometer      & 1D & Graph  & LC & \ding{55} & \ding{55} & \ding{55} \\
Che \cite{che_visualinertialwheel_2025}         & GNSS-camera-IMU-odometer-barometer & 2D & Graph & TC & \ding{55} & \ding{55} & \ding{55} \\
Ours                & GNSS-IMU-odometer      & 2D & Graph  & TC & Dataset   & \ding{51} & \ding{51} \\
\bottomrule
\end{tabular}
\end{table*}

\textit{\textbf{GNSS provides driftless measurements but affected by outliers in degraded environments: }}
Global Navigation Satellite System (GNSS) provides globally consistent, drift-free position measurements, making it a natural complement for IMU-odometer calibration in outdoor environments \cite{gao_odometer_2018}. Considering the number of wheel encoders, the GNSS-aided IMU-odometer calibration works also include 1D and 2D cases. To the best of our knowledge, \cite{wu_self-calibration_2010} is among the earliest works that utilize GPS measurements to aid IMU-odometer misalignment calibration. We call the GNSS positioning aided calibration as loose coupling (LC) and there are still some limitations: (1) reliance on sufficient satellites; and (2) inaccurate uncertainty of GNSS  positioning. A potential to solve the problems is the use of raw GNSS measurements. To overcome the limitations of LC, fuse the raw GNSS measurements in a tightly coupled manner. Recently, raw GNSS measurements are applied in multi-sensor fusion localization. Compared with the former LC systems, the raw GNSS measurements aided methods are termed as tight coupling (TC). TC approaches integrate raw GNSS such as pseudo-range and Doppler measurements have been used to couple with a LiDAR \cite{liu_glio_2024}, a camera \cite{cao_gvins_2021} and an ultra wide band (UWB) \cite{wang_tightly-coupled_2016}. In these papers, pseudo-range measurements provide the globally consistent range at meter level accuracy and Doppler measurements provide the range rate at decimeter per second level accuracy. To eliminate the effect of raw GNSS pseudo-range outliers, carrier-phase measurements with ambiguity variables are also used as range constraints at centimeter level accuracy. In recent years, the carrier-phase measurement is fused with the odometer for positioning \cite{gao_odometer_2018} and calibration \cite{che_visualinertialwheel_2025}. The results show a superior performance over the methods only using pseudo-range. 
% Chen et.al, employs differential GNSS measurements in a graph optimization framework including GNSS, Inertial, Visual, Wheel and Barometer sensors \cite{che_visual_2025}.
% However, this work has not discussed ambiguity resolution and observability. 
% On the other hand, some outlier mitigation methods also developed to improve the robustness.
% In recent years, the research trend in the past decade includes two main groups. The first group is to employ graph optimization instead of a Kalman filter for a robust estimation. For example, the authors in \cite{bai_improved_2021, bai_enhanced_2023, lyu_factor_2023} employ factor graph optimization for GNSS-INS-odometer integrated localization and calibration. 
However, there are also some limitations in current works: 1) the impact of carrier-phase ambiguity resolution in a GNSS-aided IMU-odometer calibration system is not discussed; 2) observability analysis is lacked; 3) there are no open-sourced dataset including both base/rover station GNSS measurements and IMU-odometer measurements. These limitations lead to the second gap.

These gaps motivate the development of a robust GNSS-aided IMU-odometer calibration framework that can reliably operate in degraded environments, handle raw GNSS measurements with ambiguity resolution, and provide a benchmark dataset for the community.
Therefore, our major objective is to fill the mentioned two groups of gaps. To achieve this goal, this paper proposes a GNSS-aided INS-odometer online calibration method employing raw GNSS carrier phase measurements and perform analysis on both observability and AR. The comparison of our proposed method and existing methods is illustrated in TABLE \ref{tab:related_works}. Our main contributions are as follows:

\begin{itemize}
\item \textbf{Utilize raw GNSS measurements to aid IMU-odometer online calibration with outlier mitigation and ambiguity resolution}: The integration method is based on an extendable factor graph optimization (FGO) framework \cite{wen_towards_2021}. In this FGO, the navigation state and IMU-odometer extrinsic parameters are estimated jointly, in which GNSS raw measurements including  pseudo-range, carrier-phase and Doppler are used. To the best of our knowledge, this paper is the first to employ these GNSS raw measurements together with outlier mitigation and ambiguity resolution to calibrate the IMU-odometer extrinsic and odometer's intrinsic parameters. 
\item \textbf{Derive the observability of the online calibration}: Observability analysis is conducted to show that with general motion two horizontal translation and three rotation parameters between IMU-odometer are observable, and only the vertical translation parameter at the body frame is unobservable. 
% We also verify that incorrect vertical translation at body frame does not affect AGV localization if odometer measurements are employed under a planar motion assumption.
\item \textbf{Experimental verification and open-sourced datasets}: Experiments are conducted with both simulated and real-world datasets. Ablation tests are also investigated to evaluate the effect of the GNSS-IMU lever-arm in the system. The proposed calibration method outperforms the loosely coupled methods using RTK results and shows superior accuracy and robustness in various GNSS-degraded environments. Also, this is the first paper to open source dataset including IMU, 2D-odometer and raw GNSS measurements from both rover and base stations.
\end{itemize}

The rest of the paper is organized as follows. Sec \ref{sec:overview} to Sec. \ref{sec:fgo} introduce the tightly-coupled factor graph framework, including the framework overview in Sec. \ref{sec:overview}, key sensor modeling in Sec. \ref{sec:sensor_modeling} and factor graph structure in Sec. \ref{sec:fgo}. Afterwards, Sec. \ref{sec:observability} derives the observability analysis. Then Sec. \ref{sec:simulation} and Sec. \ref{sec:field} evaluate the performance by simulation and real-world experiments. Finally, conclusions are drawn, together with an outlook in Sec. \ref{sec:conclusion}.

\section{\label{sec:overview}GNSS-constrained IMU-odometer parameter calibration Overview}
The proposed IMU-odometer parameters calibration method is based on a GNSS-INS-odometer tightly coupled integrated estimator via a nonlinear factor-graph optimization pipeline, which is designed for a differential-drive based robot car using GNSS receivers (both rover and reference stations), an IMU and an 2D odometer. The framework of the pipeline consists of two parts: (1) sensor modeling of GNSS-IMU-odometer using sensor inputs to eliminate systematic noises of raw measurements; (2) factor graph optimization based on the sensor modeling to estimate the optimal calibration parameters. 
The symbols frequently used in this paper are also introduced to define the mathematic models clearly. We present the multi-sensor coordinate frame denotation as follows:

\begin{itemize}
\item $w$: the local world frame fixed to the origin and direction of the first IMU pose (i.e., position and attitude) \cite{cao_gvins_2021};
\item $n$: the local navigation frame with the same origin as $w$ frame and east-north-up (ENU) direction \cite{shen_advancing_2024};
\item $e$: the global Earth-Centered, Earth-Fixed (ECEF) frame of GNSS fixed to the Earth \cite{li_high-precision_2023};
\item $b$: the moving body frame with the origin fixed to IMU center and sensor defined direction \cite{forster_-manifold_2017};
\item $m$: the moving vehicle-mounted frame with the origin fixed to the vehicle center and right-forward-up (RFU) direction .
\end{itemize}

 Additionally, the transformation from $a$ to $b$ with $a$ as the reference frame is denoted as translation $p^a_b$  and rotation $R^a_b$ separately. In addition, the skew-symmetric operator is defined as: 
\begin{equation}
     \left\lfloor \mathbf{a}\times \right\rfloor = 
\begin{bmatrix}
0 & -a_3 & a_2 \\
a_3 & 0 & -a_1 \\
-a_2 & a_1 & 0 
\end{bmatrix} 
 \end{equation}
where $\mathbf{a}=\begin{bmatrix}
a_1 &
a_2 &
a_3
\end{bmatrix}^T \in \mathbb{R}^3$.  
\section{\label{sec:sensor_modeling}Sensor Modeling}
In this section, we describe in detail the sensor modeling based on raw sensor measurement inputs. In terms of inputs for sensor modeling, GNSS rover and reference stations provide broadcast ephemeris and raw measurements such as pseudo-range, carrier-phase and Doppler measurements; IMU outputs linear accelerations and angular velocities; odometer offers forward linear velocity and bearing angular velocity \cite{de_giorgi_online_2023}. This paper adopts the same IMU model as the one in \cite{forster_-manifold_2017}, hence only GNSS and odometer models are introduced as below while the details of the IMU model can be referred to \cite{forster_-manifold_2017} and \cite{chi_gici-lib_2023}. 

\subsection{GNSS Modeling}

GNSS raw measurement models including pseudo-range $P$,  carrier-phase $L$ and Doppler shift $D$ between the satellite $s$ and receiver $r$ can be formulated as \cite{li_review_2022}:
\begin{equation}
\left\{    
\renewcommand{\arraystretch}{1.5}
    \begin{array}{l}
    P = \rho + I + T + c(t_r - t^s) + c(b_r - b^s) + \epsilon_P, \\
    L = \rho - I + T + c(t_r - t^s) + c\lambda(B_r - B^s) + \lambda N + \epsilon_L, \\
    \lambda D = \frac{\Delta \mathbf{p}^T \Delta \mathbf{v}}{\rho} + c(\dot{t_r} - \dot{t^s}) + \epsilon_D.
    \end{array}\label{con:gnss_raw}
\right.
\end{equation}
where $\rho$ is the geometric distance between the phase center of the satellite and receiver antennas; $\Delta \mathbf{p}$ and $\mathbf{v}$ are the relative position and velocity between the satellite and receiver antenna phase center ($\rho=\left\| \Delta \mathbf{p} \right\| $); $I$ and $T$ denote the ionosphere and troposphere delays; $c$ is the speed of light; $t_r$ and $t^s$ are the clock offsets of the receiver and satellite; $\dot{t_r}$ and $\dot{t^s}$ are the clock shift rate of the receiver and satellite; $b_r$ and $b^s$ are the receiver and satellite code hardware bias; $B_r$ and $B^s$ are the receiver and satellite phase bias; $\lambda$ and $N$ are the wavelength and integer ambiguity of carrier signal; $\epsilon_P$, $\epsilon_L$ and $\epsilon_D$ represent the sum noise and multipath error of the corresponding measurements.

In view of the systematic errors in GNSS raw measurements, the double-differenced (DD) model is usually adopted to eliminate most of systematic errors in pseudo-range $P$ and carrier-phase $L$ \cite{wen_towards_2021}. Specifically, the DD model includes single-differenced (SD) model between satellites and receivers: SD model between satellites can eliminate the receiver-related bias; the SD model between receivers can eliminate the satellite-related bias; the DD model can greatly reduce the atmospheric delays. After applying DD model, the GNSS DD measurements can be expressed as:
\begin{equation}
\left\{
    \renewcommand{\arraystretch}{1.5}
    \begin{array}{l}
    P_{DD} = \rho_{DD} + \epsilon_{P_{DD}} , \\
    L_{DD} = \rho_{DD}  + \lambda_{DD} {N_{DD}}  + \epsilon_{L_{DD}} .
    \end{array}
\right. \label{equ:dd_combine}
\end{equation}
where $P_{DD}$ and $L_{DD}$ are the DD pseudo-range and DD carrier-phase; $\rho_{DD} $ is the DD geometric distance between satellite and receiver antenna phase center; $\lambda_{DD}$ and $N_{DD}$ are the wavelength and ambiguity of DD carrier-phase; $\epsilon_{P_{DD}}$ and $\epsilon_{L_{DD}}$ represent the noise and residual bias of the corresponding measurements.

Notably, although the DD operator eliminates most systematic errors within the GNSS measurements, the unmodeled noises and outliers including multi-path and NLOS effects are amplified. Therefore, in this work, we also employ our previously proposed two-stage GNSS outlier detection method \cite{song_two_2025}, which first leverages Doppler measurements to detect pseudo-range outliers and then uses pre-integrated IMU and odometer constraints to refine the rejection of remaining errors, resulting in significantly improved positioning accuracy and robustness.

\subsection{Odometer Modeling}
Different from existing work \cite{bai_improved_2021} for autonomous driving, we focus on a general odometer model for ground robots. One of the general models is the differential drive model which can provide planar kinematic constraints provided that measurements between time $k$ and $k+1$ are collected on a supporting planar manifold \cite{he_camera-odometer_2017}. According to the differential drive model, we model the true odometer output fixed to the $m$ frame as:
\begin{equation}
\left\{
    \renewcommand{\arraystretch}{1.5}
    \begin{array}{l}
        {v}^m = (1+s_v) (\hat{v}^m-\epsilon_{v^m}), \\ 
        {\omega}^m = (1+s_\omega)(\hat{\omega}^m-\epsilon_{\omega^m})
    \end{array}
\right. \label{con:odo_raw}
\end{equation}
where ${v}^m$ and ${\omega}^m$ are the true linear velocity in the forward direction and angular velocity in the up direction separately; $s_v$ and $s_\omega$ are the scaling factors of linear velocity and angular velocity from the odometer; $\hat{v}^m$ and $\hat{\omega}^m$ present the forward linear velocity and bearing angular velocity measurements of the odometer; $\epsilon_{v^m}$ and $\epsilon_{\omega^m}$ are the corresponding measuring noises. The odometer scaling factors are modeled as random walk \cite{thrun_probabilistic_2006}: $\dot{s_v}=\epsilon_{s_v}$ and $
\dot{s_\omega}=\epsilon_{s_\omega}$. 

Under the planar motion assumption,  the odometer measurements can be projected to  $\mathbf{v}^m \in \mathbb{R}^3$ and $\mathbf{\omega}^m \in \mathbb{R}^3$ in three dimensions. The projected odometer measuring model can be expressed as:
\begin{equation}
\begin{bmatrix}
 {\mathbf{v}^m}^T &
 {\mathbf{\omega}^m}^T
\end{bmatrix}^T = \mathcal P
\begin{bmatrix}
 {v}^m & {\omega}^m 
\end{bmatrix}^T.
\label{equ:projection}
\end{equation}

where $\mathcal P=\begin{bmatrix}
 \mathbf{0}&  \mathbf{e}^T_2&\mathbf{0}  &\mathbf{0}  &\mathbf{0}  &\mathbf{e}^T_3
\end{bmatrix}^T$ is the projection matrix with $\mathbf{e_i}$ being the $3\times 1$ unit vectors with the $i$th element of 1.

\section{\label{sec:fgo}Factor Graph Optimization}
In this section, we introduce the factor graph optimization framework based on the sensor modeling section.

The proposed factor graph framework is shown in Fig. \ref{fig:factor_graph_structure}, containing two kinds of nodes-variable nodes and factor nodes. Variable nodes denote the states needed to be estimated and factor nodes represent the sensor observation constraints related to the states \cite{bai_enhanced_2023}. All the nodes are formulated in a sliding window for numerical programming by a nonlinear estimator after which the estimator will output the navigation state, and use the latest estimated IMU-odometer parameters for sensor calibration at the next factor graph estimating moment. The following section firstly describes the state equations and then derives multiple sensor constraints, including GNSS, IMU and odometer constraints. To simplify the expression, we use  $\mathcal{G}, $ $\mathcal{I}$ and $\mathcal{O}$ to denote GNSS , IMU and odometer, respectively. 

\subsection{State Formulation}
To begin with, we define the complete states vector $\mathbf{X}$ including multiple variable nodes $\mathbf{x}_k$within a sliding window as:
\begin{figure}
    \centering
    \includegraphics[width=1\linewidth]{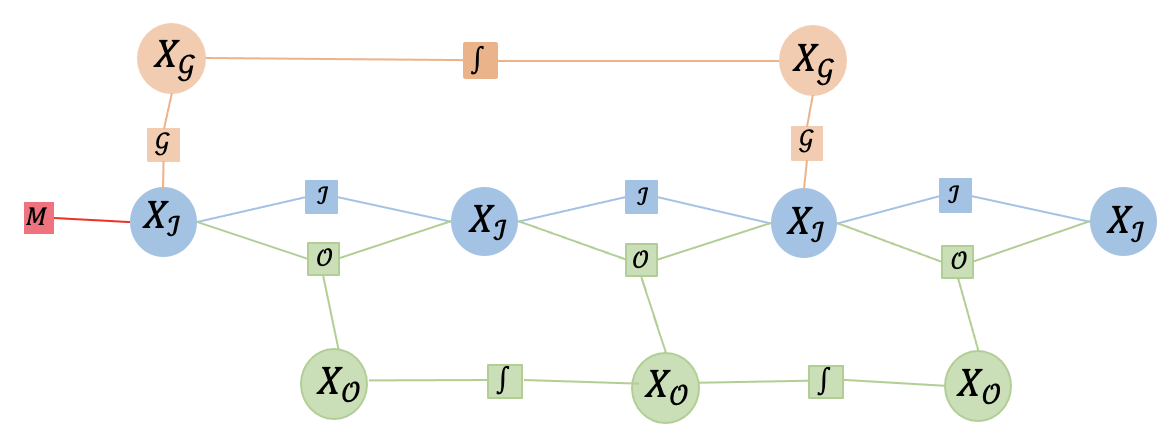}
    \caption{The proposed factor graph with a sliding window overview, where the red box denotes the marginalization factor, the orange, blue, and green boxes denote GNSS, IMU, odometer measuring and relative state constraints, and the orange, blue, and green circles denote GNSS, IMU, and odometer states.}
    \label{fig:factor_graph_structure}
\end{figure}

\begin{equation}
    \mathbf{X}=[\mathbf{x}_1^T,...,\mathbf{x}_k^T,...,\mathbf{x}_n^T]^T
\end{equation}
where $\mathbf{X}$ denotes the robot state set with $n$ state in the sliding window, and  $\mathbf{x}_k$ signifies the state at epoch $k$ consisting of the IMU state $\mathbf{x}_{\mathcal{I}}$, GNSS state $\mathbf{x}_{\mathcal{G}}$ and odometer state $\mathbf{x}_{\mathcal{O}}$:
\begin{equation}
\mathbf{x}_k=[\mathbf{x}_{\mathcal{I},k}^T, \mathbf{x}_{\mathcal{G},k}^T, \mathbf{x}_{\mathcal{O},k}^T]^T
\end{equation}
Although $\mathbf{x}_{\mathcal{G}}$ and $\mathbf{x}_{\mathcal{O}}$ will not be added to the same state due to asynchronous measuring of GNSS and odometer. For presentation simplicity, we describe them explicitly in the same state vector. Specifically, the INS states can be written as:
\begin{equation}
\mathbf{x}_{\mathcal{I},k}=[{\mathbf{p}^w_{b,k}}^T, {\mathbf{v}^w_{b,k}}^T, {\mathbf{q}^w_{b,k}}^T, \mathbf{b}_{a,k}^T, \mathbf{b}_{g,k}^T]^T
\end{equation}
where ${\mathbf{p}^w_{b}}$ is the robot position in the world frame; ${\mathbf{v}^w_{b}}$ is the robot velocity; ${\mathbf{q}^w_{b}}$ is the robot attitude quaternion; $\mathbf{b}_{a}$ and $\mathbf{b}_{g}$ are the bias of the accelerometer and gyroscope. For the GNSS states, considering the convenience of selecting new reference satellite, we keep the single-differenced ambiguity set $\mathbf{N}_{SD,k}$ between two receiver stations together with the receiver clock shift rate $\dot{t}_{r,k}$ in the state vector (because of $D$ is not single differenced) as below: 
\begin{equation}
\mathbf{x}_{\mathcal{G},k}=[\dot{t}_{r,k}, \mathbf{N}_{SD,k}]
\end{equation}
For the odometer state, both of odometer scaling factors and IMU-odometer extrinsic transformation (translation $\mathbf{p}_m^b$ and rotation $\mathbf{q}_m^b$) are included. Note that new scaling factors are inserted at every odometer update moment and there is only one IMU-odometer extrinsic transformation in the sliding window.
\begin{equation}
\mathbf{x}_{\mathcal{O},k}=[s_{v_k}, s_{\omega_k}, {\mathbf{p}_m^b}^T, {\mathbf{q}_{m}^b}^T]^T
\label{equ:odometer_state}
\end{equation}

The cost function $r$ to be minimized includes the combination of sensor measurement constraints is formulated as:
\begin{equation}
    r = r_\mathcal{M} + r_\mathcal{G} + r_\mathcal{I} + r_\mathcal{O}  
\end{equation}
where $r_\mathcal{M}$ denotes the prior marginalization constraint; $r_\mathcal{G}$ denotes the GNSS constraint; $r_\mathcal{I}$ and $r_\mathcal{G}$ represent the inertial constraint and the odometer constraint, both of which are constructed by measurement integration. Due to page limit, we omit the details of $r_\mathcal{M}$ and readers are encouraged to refer to \cite{ando_generalized_1979}. The rest constraints are introduced as below.

\subsection{GNSS Constraints}
Since we define the position fixed to IMU $b$ frame in the state, a lever arm $\mathbf{p}^g_b$ between IMU and GNSS is utilized to associate the GNSS antenna position $\mathbf{p}^e_g$  and velocity $\mathbf{v}^e_g$ with the IMU-center position $\mathbf{p}^w_b$ and velocity $\mathbf{v}^w_b$. Specifically, considering the lever arm between IMU and GNSS antenna is constant, the GNSS antenna position used in \eqref{equ:dd_combine} is:
\begin{equation}
\left\{
    \renewcommand{\arraystretch}{1.5}
    \begin{array}{l}
        \mathbf{p}^e_g = \mathbf{R}^e_n \mathbf{R}^n_w(\mathbf{p}^w_{b}+\mathbf{R}^w_{b}\cdot \mathbf{p}^b_g), \\
        \mathbf{v}^e_g = \mathbf{R}^e_n \mathbf{R}^n_w(\mathbf{v}^w_{b}+\left\lfloor \mathbf{\omega}_{ib}^b\times \right\rfloor\cdot \mathbf{p}^b_g) .
    \end{array}
\right. \label{con:gnss_imu_trans}
\end{equation}

Based on the GNSS model \eqref{con:gnss_raw} \eqref{equ:dd_combine} and state transforming model \eqref{con:gnss_imu_trans}, the cost function of GNSS could be generically written in the following form:
\begin{equation}
    r_{R}= \left\|  P_{DD} -  \hat{P}_{DD}) \right\|^2_{\Sigma_{P}}  +  \left\|  L_{DD} -  \hat{L}_{DD}) \right\|^2_{\Sigma_{L}}  +  \left\|  D_{} -  \hat{D}_{}) \right\|^2_{\Sigma_{D}}
\end{equation}
where $\Sigma_{P}$, $\Sigma_{L}$ and $\Sigma_{D}$ present the covariance matrix of corresponding measurements computed by the elevation angle based model \cite{chi_gici-lib_2023}.
Moreover, we utilize the LAMBDA method to perform ambiguity resolution in this paper \cite{teunissen_lambda_2006}. If the DD ambiguity resolution succeeds, the integer ambiguity constraints will also be employed as a factor in the later sliding window:
\begin{equation}
    r_{AR}= \left\|  \hat{N}_{DD,ij} - g( {N}_{SD}^i-{N}_{SD}^j)\right\|^2_{\Sigma_{AR}}
\end{equation}
where $\hat{N}_{DD,ij}$ denotes the integer double differenced ambiguity estimated by the LAMBDA; ${N}_{SD}^i$  denotes the single differenced integer ambiguity state between rover and base station for satellite $i$; function $g$ denotes the single differenced model between satellite $i$ and $j$; $\Sigma_{AR}$ denotes the covariance matrix of the integer ambiguity and it is defined as a diagonal matrix with the element $1\times10^{-6}$ in this paper.

Hence, the residual model of GNSS can be presented as the sum of them:
\begin{equation}
    r_\mathcal{G}=   r_{R} + r_{AR}
\end{equation}

\subsection{Inertial Constraints}

To deal with high-rate IMU measurements, a pre-integration model is employed to provide relative motion constraints between two consecutive epochs. Plenty of existing works have introduced the pre-integration model of IMU in details and comprehensive definition and derivation can be found in \cite{forster_-manifold_2017}. In short, the IMU pre-integration constraint factor provides the relative connection between two states $\mathbf{x}_{k-1}$ and $\mathbf{x}_{k}$, which can be written as:
\begin{equation}
    r_{IMU}= \left\| \mathbf{x}_k \boxminus f(\mathbf{x}_{k-1},\mathcal{I}_{a}, \mathcal{I}_{g}) \right\|^2_{\Sigma_\mathcal{I}}
\end{equation}
where $\boxminus$ denotes the generalized minus on manifold defined in \cite{forster_-manifold_2017}; $\mathcal{I}_{a}$ and  $\mathcal{I}_{g}$ denote the IMU accelerometer and gyroscope measurements between epoch $k-1$ and $k$;  $\Sigma_\mathcal{I}$ represents the covariance matrix of the integral measurements, which is calculated by the same model in \cite{chi_gici-lib_2023}. In addition, we also use the IMU measurements to detect the motion of the AGV and apply motion constraints in \cite{chi_gici-lib_2023}, such as  zero-velocity update (ZUPT), non-holonomic constraint (NHC) and heading measurement constraint (HMC), to enhance the state estimation. Therefore, the inertial constraint can be written as:
\begin{equation}
    r_\mathcal{I}=r_{IMU}+r_{motion}
\end{equation}

\subsection{Odometer constraints}
Based on the projected odometer measuring model \eqref{equ:projection} and state model  \eqref{equ:odometer_state}, we can transform the odometer measurements from $m$-frame to $b$-frame with IMU-odometer extrinsic transformation. The odometer velocity in $b$-frame can be expressed by: 
\begin{equation}
\left\{
    \renewcommand{\arraystretch}{1.5}
    \begin{array}{l}
        \mathbf{v}^b = \mathbf{R}_m^b\mathbf{v}^m -\left\lfloor \mathbf{\omega}^b_{ib}\times \right\rfloor \mathbf{p}_m^b, \\
        \mathbf{\omega}^b = \mathbf{R}_m^b \mathbf{\omega}^m.
    \end{array}
\right. \label{con:odo_raw}
\end{equation}
Then similar to the IMU constraints, the integral measurements in $b$-frame between two epochs $k$ and $k+1$ can be written as the equations \eqref{equ:integral_odo}, where $\delta \mathbf{p}^{k}_{\mathcal{O},k+1}$ and $\delta \mathbf{q}^{k}_{\mathcal{O},k+1}$ denote integral odometer position and attitude measurements; $\Omega'$ denote the quaternion right-hand multiplier in \cite{forster_-manifold_2017}.

\begin{figure*}
\begin{equation}
\left\{
    \renewcommand{\arraystretch}{1.5}
    \begin{array}{l}
        \mathbf{R}_w^{b_k}\cdot \mathbf{p}_{b_{k+1}}^w = \mathbf{R}_w^{b_k}\mathbf{p}_{b_{k}}^w + \underbrace{\int_{k}^{k+1}\mathbf{R}^{b_k}_{b_t}[\mathbf{R}^b_m\mathbf{e}_2 (1+s_v) \hat{v}^m-\left\lfloor \mathbf{\omega}^b_{ib}\times \right\rfloor\mathbf{p}_m^b]\cdot dt}_{\delta \mathbf{p}^{k}_{\mathcal{O},k+1}}, \\
        \mathbf{q}_w^{b_k}\otimes \mathbf{q}_{b_{k+1}}^w = \underbrace{\int_{k}^{k+1}\frac{1}{2} \Omega'[\mathbf{R}_m^b\mathbf{e}_3 (1+s_\omega) \hat{\mathbf{\omega}}^m]\mathbf{q}_{b_t}^{b_k}\cdot dt}_{\delta \mathbf{q}^{k}_{\mathcal{O},k+1}}.
    \end{array}
\right. \label{equ:integral_odo}
\end{equation}
\end{figure*}

\begin{figure*}
\begin{equation}
F_t = \begin{bmatrix}
 \mathbf{0}&  -\left\lfloor \mathbf{R}^{b_k}_{b_t}(\mathbf{R}^b_m\mathbf{e}_2 (1+s_v) \hat{v}^m_t-\left\lfloor \mathbf{\omega}^b_{ib}\times \right\rfloor\mathbf{p}_m^b)\times \right\rfloor &  \mathbf{R}^{b_k}_{b_t}\mathbf{R}^b_m\mathbf{e}_2\hat{v}^m_t&  \mathbf{0}\\
 \mathbf{0}&   -\left\lfloor \mathbf{R}^b_{m}\mathbf{e}_3 (1+s_\omega) \hat{\omega}^m_t\times \right\rfloor&  \mathbf{0}&  \mathbf{R}^b_{m}\mathbf{e}_3\hat{\omega}^m_t  \\
 \mathbf{0}&  \mathbf{0}&  0&  0  \\
 \mathbf{0}&  \mathbf{0}&  0&  0 
\end{bmatrix}  
\label{equ:F_t}
\end{equation}

\end{figure*}

To derive the uncertainty of the integral measurements, continuous error propagation model of the integral odometer measurements is given as: 

\begin{equation}
\begin{bmatrix}
 \delta \dot{\mathbf{p}}^{k}_{\mathcal{O},t}\\
 \delta \dot{\mathbf{q}}^{k}_{\mathcal{O},t}\\
 \delta \dot{s}_{v,t}\\
 \delta \dot{s}_{\omega,t}
\end{bmatrix} = \mathbf{F_t}\cdot 
\begin{bmatrix}
 \delta \mathbf{p}^{k}_{\mathcal{O},t}\\
 \delta \mathbf{q}^{k}_{\mathcal{O},t}\\
 \delta s_{v,t}\\
 \delta s_{\omega,t}
\end{bmatrix}
+  \mathbf{G_t}\cdot
\begin{bmatrix}
 \epsilon_{v^m}\\
 \epsilon_{\omega^m}\\
 \epsilon_{s_{v}}\\
 \epsilon_{s_{\omega}}
\end{bmatrix}\label{equ:odo_propagate}
\end{equation}

The details of the Jacobian matrix $ \mathbf{F}_t$ and $ \mathbf{G}_t$ in the uncertainty propagation equation are given as in \eqref{equ:F_t} and \eqref{equ:G_t}.
\begin{equation}  
  \resizebox{0.8\hsize}{!}{$\begin{aligned}
 G_t=\begin{bmatrix}
 -\mathbf{R}^{b_k}_{b_t}\mathbf{R}^b_m\mathbf{e}_2 (1+s_v) &\mathbf{0}   &\mathbf{0}  &\mathbf{0}  \\
 \mathbf{0}&  -\mathbf{R}^b_{m}\mathbf{e}_3 (1+s_\omega)&\mathbf{0}&\mathbf{0}  \\
 \mathbf{0}&  \mathbf{0}&  1&0  \\
\mathbf{0}&  \mathbf{0}&  0&1 
\end{bmatrix}
  \end{aligned}$}\label{equ:G_t}
\end{equation}

Based on the integral odometer measurements, we derive the cost function of odometer kinematic in the form:
\begin{equation}
    r_\mathcal{O}= \left\| x_k \boxminus f(x_{k-1},\mathcal{O}_{v}, \mathcal{O}_{\omega}) \right\|^2_{\Sigma_\mathcal{O}}
\end{equation}
where $\mathcal{O}_{v}$ and  $\mathcal{O}_{\omega}$ denote the odometer measurements between epoch $k-1$ and $k$; $\Sigma_\mathcal{O}$ represents the covariance matrix of the integral measurements, which is given by (\ref{equ:odo_propagate}).

\section{\label{sec:observability}Observability Analysis}
In addition to the FGO framework, observability analysis for specific parameters is derived in this section. This is because the calibration method utilizes projected 3D kinematic measurements from differential-drive based odometer, so it will be interesting to analyze the local observability of the odometer scaling factor and transformation parameter between IMU and the vehicle with the raw planar odometer constraints. Since the local observability of scaling factors is derived in \cite{vasilyuk_identification_2020}, we focus on observability of extrinsic transformation parameters following the method proposed in \cite{zuo_visual-inertial_2019}. To simplify the process, we assume that the IMU bias is estimated accurately with GNSS constraints and the system is able to provide virtual linear and angular velocity measurements in the IMU's body frame:
\begin{equation}
    \breve{\mathbf{v}}^b = {\mathbf{v}}^b + \epsilon_{v^b},  
    \breve{\mathbf{\omega}}^b = {\mathbf{\omega}}^b + \epsilon_{\omega^b}
    \label{body_vel}
\end{equation}
where $\breve{\mathbf{v}}^b$ and $\breve{\mathbf{\omega}}^b$ are the virtual measurements from the GNSS/IMU integration. Here we introduce a lemma about the observability.

\textbf{Lemma 1.} \textit{With the assumption of planar motion, the translation parameter along the vehicle rotation axis is not locally identifiable, while translation and rotation extrinsic parameters along other directions are locally identifiable under general motion.}

\textit{Proof.} Substituting \eqref{body_vel} into \eqref{con:odo_raw} leads to 
\begin{equation}
\begin{bmatrix}
 \breve{r}_v\\
 \breve{r}_\omega 
\end{bmatrix}
    =
\begin{bmatrix}
[(\mathbf{R}^m_b(\breve{\mathbf{v}}^b + \left\lfloor \breve{\mathbf{\omega}}^b\times \right\rfloor \mathbf{p}_m^b) - \mathbf{v}^m) \\
\mathbf{R}_b^m \breve{\mathbf{\omega}}^b-\mathbf{\omega}^m
\end{bmatrix}
\label{con:odo_analysis}
\end{equation}

Note that the equation only contains virtual measurements and the IMU-odometer transformation parameters. A necessary and sufficient condition of the IMU-odometer transformation $[{\mathbf{p}_m^ b}^T,{\mathbf{q}_{m}^b}^T]^T$to be locally identifiable is that the following observability matrix $\mathbf{M}$ has full column rank:
\begin{equation}
    \mathbf{M} = 
\begin{bmatrix}
 \mathbf{M}(t_0)^T&\mathbf{M}(t_1)^T& ... &\mathbf{M}(t_s)^T
\end{bmatrix}^T
\end{equation}
where
\begin{equation}
    \mathbf{M}(t) 
=
\begin{bmatrix}
\mathbf{R}^m_b \left\lfloor \breve{\mathbf{\omega}}^b\times \right\rfloor&
\mathbf{R}^m_b(\mathbf{v}^b + \left\lfloor \breve{\mathbf{\omega}}^b\times \right\rfloor \mathbf{p}_m^b)\times \\
 \mathbf{0}&\mathbf{R}_b^m \breve{\mathbf{\omega}}^b\times 
\end{bmatrix}
\end{equation}
A necessary and sufficient condition of the extrinsic transformation to be locally identifiable is that $\mathbf{M}$ has full column rank. However, if the robot car is operating on a plane, the third column of  $\mathbf{R}^m_b \left\lfloor \breve{\mathbf{\omega}}^b\times \right\rfloor$ is zero, which demonstrates that the translation parameter along the Z-axis of $m$-frame is not identifiable. For other parameters, if the robot can provide sufficient acceleration and rotation, they will be locally identifiable. This completes the proof.

\begin{algorithm}
  \caption{FGO-based Raw GNSS Aided IMU-Odometer Online Calibration}
  \label{alg:fgo_calib}
  \begin{algorithmic}[1]
  \REQUIRE Raw GNSS pseudo-range/carrier-phase/Doppler measurements from rover and base station GNSS receivers, IMU and odometer measurements; initial state $\mathbf{X}_0$ including IMU-odometer parameter $\mathbf{x}_{\mathcal{O},0}$ and its initial covariance $\mathbf{P}_0$
  \ENSURE refined states $\mathbf{\mathbf{\hat{X}}}$ including online calibration parameters $\mathbf{\hat{x}}_{\mathcal{O}}$, 
  
  \STATE \% Preprocess raw GNSS data, detect and reject outliers via two-stage outlier mitigation
  \STATE \% Initialize IMU-odometer parameters ($\mathbf{x}_{\mathcal{O}}$) included in state variables ($\mathbf{X}$)
  
  \FOR{each sliding window $k=1,\dots,K$}
      \STATE Formulate factor graph with inertial, odometer, GNSS factors using residual $r$ and state $\mathbf{X}$
      \STATE Perform nonlinear optimization to jointly estimate $\mathbf{\mathbf{\hat{X}}}$, resolve DD ambiguities $\hat{N}_{DD}$ using LAMBDA
      \STATE If ambiguity ratio test passes, fix integer ambiguities
      \STATE Update calibration parameters $\mathbf{\hat{x}}_{\mathcal{O}}$ online
  \ENDFOR
  
  \RETURN $\mathbf{\mathbf{\hat{X}}}$
  \end{algorithmic}
\end{algorithm}

\section{\label{sec:simulation}Simulation Analysis}
\subsection{Introduction of the Simulation}
\subsubsection{simulation setup}
To analyze the performance of the proposed calibration method, a simulation test is performed using numerically synthetic raw GNSS, IMU, and odometer measurements. With the simulation test, it is possible to set true extrinsic parameters and appropriate measuring noise to evaluate the accuracy and observability of calibration while mostly eliminating uncontrolled uncertainties like GNSS multi-path effect, IMU bias instability and wheel slip. For the simulation, a Gazebo 11.0 simulator \cite{koenig_design_2004} is employed to provide the ground-truth trajectory, which has been widely applied to support autonomous robot systems. Then, raw GNSS measurements of both rover and base stations are simulated based on the GREAT software \cite{li_great-pvt_2025} with precise ephemeris and clock products from NASA CDDIS \cite{noll_crustal_2010} and the above ground-truth trajectory from Gazebo. The IMU and odometer measurements are generated with given sensor configuration. The brief introduction of simulated sensor configuration is presented in Table \ref{tab:sim_config}. 
\begin{figure}
    \centering
    \includegraphics[width=1\linewidth]{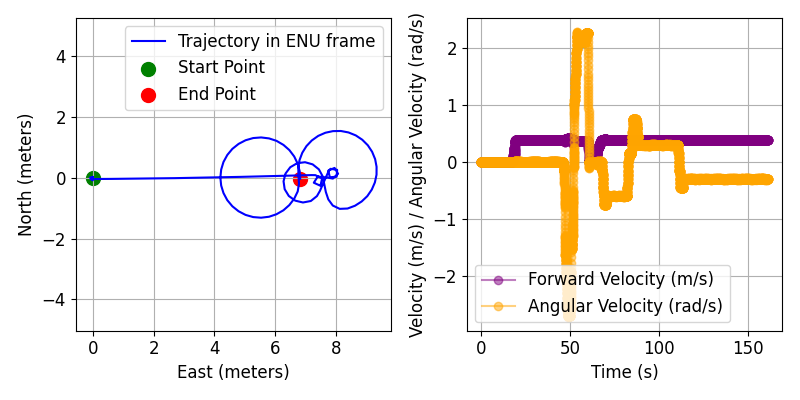}
    \caption{Simulated trajectory and odometer measurements.}
    \label{fig:sim_traj_vel}
\end{figure}

\renewcommand{\arraystretch}{1.5}
\begin{table}[]
\centering
\caption{Simulation configuration}
\label{tab:sim_config}
\begin{tabular}{p{4cm} p{4cm}}
\hline
Simulation settings                   & values                       \\ \hline
GNSS constellations                   & GPS, Galileo, BDS            \\
Satellite number                      & 19$\sim$21                   \\
Position dilution of precision (PDOP) & 1.25$\sim$1.45               \\
GNSS frequency                        & L1, L2                       \\
Pseudo-range noise        & $0.3*\sqrt(1+1/sin(El)) m$   \\
Carrier-phase noise      & $0.003*\sqrt(1+1/sin(El)) m$ \\
GNSS sampling rate                    & 1 Hz                         \\
Gyroscope bias (not employed in estimation)                        & 900 $\degree/h$              \\
Accelerometer bias (not employed in estimation)                      & 5 $mGal$                     \\
Angle random walk noise                  & 20 $\degree/\sqrt(h)$         \\
Velocity random walk noise                  & 0.1 $m/s/\sqrt(h)$           \\
IMU sampling rate                     & 100 Hz                       \\
odometer linear velocity noise        & 0.01 $m/s$                     \\
odometer angular velocity noise       & 1 $\degree/s$                \\
odometer scaling factor               & 0, 0              \\
odometer random walk noise               & 0 $\sqrt(Hz)$, 0 $\sqrt(Hz)$           \\
odometer sampling rate& 25 Hz                        \\ \hline
\end{tabular}
\end{table}

To evaluate the observability conditions for general ground robots, a typical trajectory with only horizontal motion is simulated for the AGV. The ground-truth trajectory and velocity in the body frame are illustrated in Fig. \ref{fig:sim_traj_vel}, consisting of several types of motion to test the calibration process (e.g., standing stationary, accelerating in straight lines and going around a circle). 
\begin{figure*}
        \centering
        \includegraphics[width=0.9\linewidth]{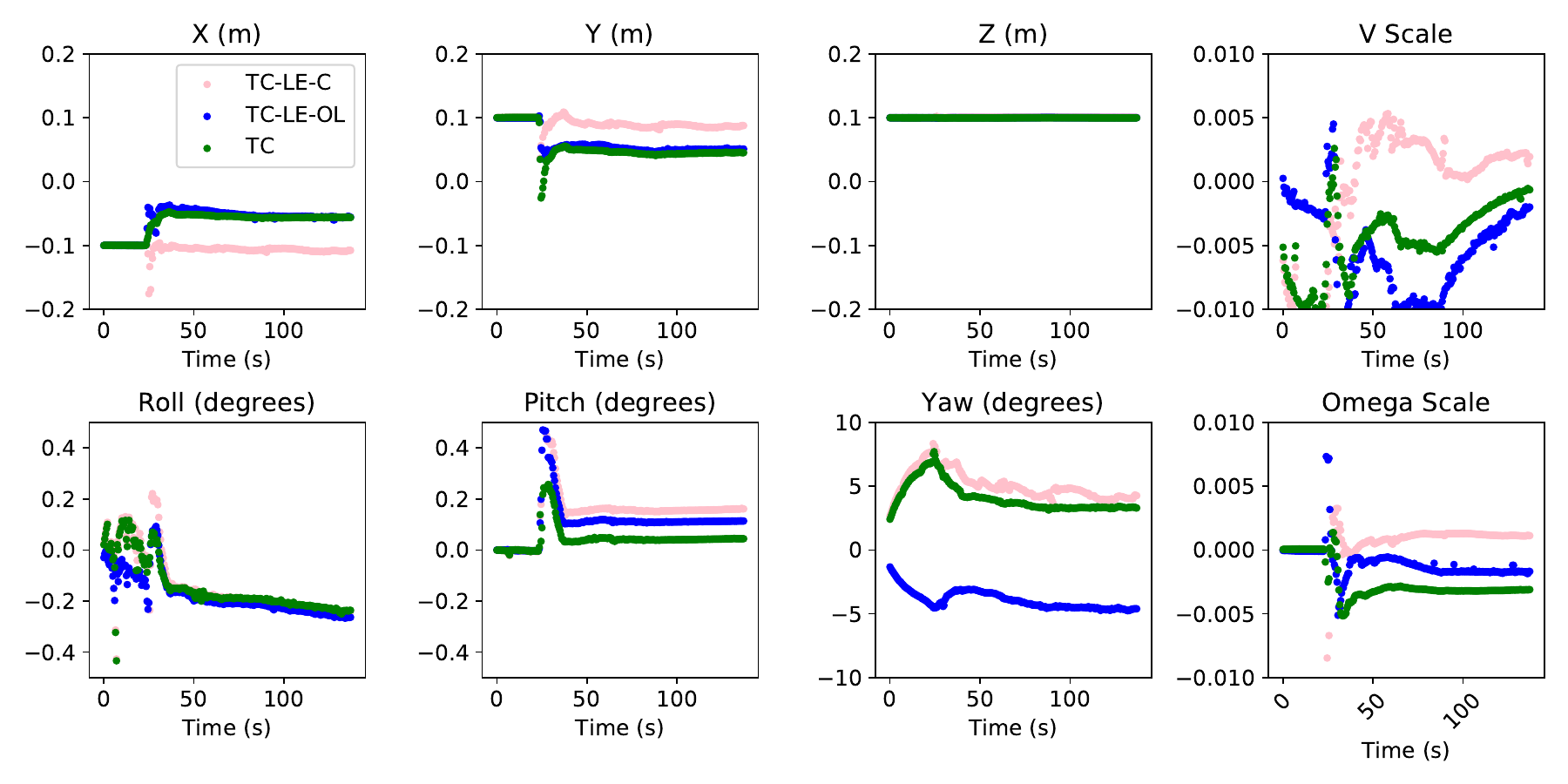}
        \caption{Calibration errors in the Simulation, where the pink line denotes the TC method with constant GNSS-IMU lever-arm error (TC-LE-C), the blue line denotes the TC method with online estimated GNSS-IMU lever-arm (TC-LE-OL), the green one denotes the TC method without GNSS lever-arm disturbance (TC).}
        \label{fig:sim_result}
\end{figure*}

\subsubsection{Evaluated Methods}
Due to the bias instability of MEMS-IMU, it is difficult for IMU-odometer fusion to calibrate the IMU-odometer parameters. Besides, current sensors used in IMU-odometer calibration methods includes camera \cite{zhou_online_2022}, GNSS \cite{bai_enhanced_2023} and other sensors \cite{lee_mins_2023} while each kind of sensor is designed for specific scenes. In this simulation, we focus on the accuracy and observability performance with ideal measuring configurations and the comparison with other methods is performed in the field test. In the following, we compare the proposed method with the ground truth to validate the calibration accuracy. At the same time, we also consider the effect of GNSS-IMU lever-arm error. The introduction of compared methods is shown as below:

(a)	TC-LE-C: the tightly coupled (TC) GNSS/IMU/odometer integration method with constant GNSS-IMU lever-arm (LE-C). The LE error is 0.1 m in each axis of $\mathit{b}$-frame.

(b)	TC-LE-OL:  the TC method with online GNSS-IMU lever-arm estimation (LE-OL). The LE error is 0.1 m in each axis of $\mathit{b}$-frame.

(c)	TC: the TC method without GNSS-IMU lever-arm errors.

\begin{table*}[t]
\centering
\caption{Mean calibration errors over 40 Monte-Carlo simulations. Translation errors are in meters, rotation errors in degrees, scaling factor errors unitless.}
\label{tab:sim_calib_errors}
\begin{tabular}{@{}lcccccccc@{}}
\toprule
Scheme & $\text{X (m)}$ & $\text{Y (m)}$ & $\text{Z (m)}$ & Roll (deg) & Pitch (deg) & Yaw (deg) & V Scale & Omega Scale\\ 
\midrule
TC        & 0.06 & 0.05 & –     & 0.24 & 0.05 & 3.29 & $6.0\times10^{-4}$ & $3.1\times10^{-3}$ \\
TC-LE-C   & 0.11 & 0.09 & –     & 0.24 & 0.16 & 4.26 & $2.0\times10^{-3}$ & $1.1\times10^{-4}$ \\
TC-LE-OL  & 0.06 & 0.05 & –     & 0.26 & 0.11 & 4.60 & $2.0\times10^{-3}$ & $3.1\times10^{-3}$ \\
\bottomrule
\end{tabular}
\end{table*}

\begin{figure*}
    \centering
    \includegraphics[width=1\linewidth]{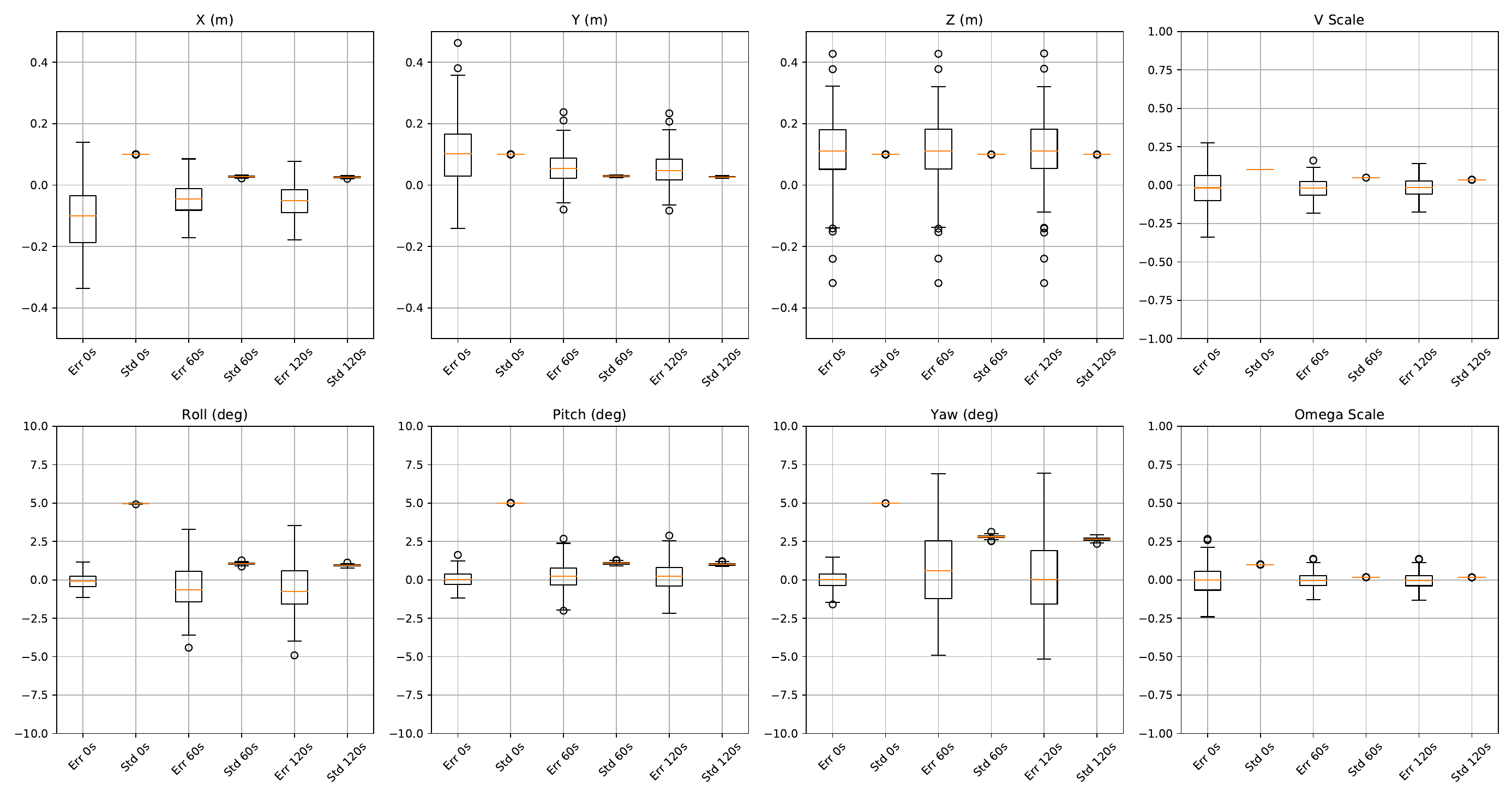}
    \caption{Statistics for calibration results and uncertainty in 100 Monte-Carlo simulation with random initial value. "Err 0s", "Err 60s" and "Err 120s" mean the calibration error at corresponding epochs. "Std 0s", "Std 60s" and "Std 120s" mean the standard deviation of these calibrated results.}
    \label{fig:sim_boxplot}
\end{figure*}

\subsection{Calibration results in simulation tests}\label{sec:sim_result}
We first compare the calibration results from the methods above in Fig. \ref{fig:sim_result}, including errors of IMU-odometer extrinsic transformation and odometer scaling factors. At 20 s, the AGV quickly raise its forward velocity to 0.25 m/s and keep constant velocity in about 20 s. Since the velocity in most of this period is constant. Most of the translation and rotation extrinsic parameters hardly change except yaw angle and scaling for v ($s_v$). At 45 s, the AGV starts to circle around and most of the parameters converge in seconds. In the following time, all the extrinsic parameters keep stable until the end of the simulation test. 
After the calibration, the absolute calibration errors at final epochs are concluded in TABLE \ref{tab:sim_calib_errors}. Notably, translation error for Z-axis of $m$-frame does not change during the 2D motion. This verifies that the vertical translation at $m$-frame is unidentifiable, so we do not analyze its error below. For scheme TC, the translation errors are 0.06 m for X-axis and 0.05 m for Y-axis. The rotation errors are 0.24$\degree$, 0.05$\degree$ and 3.29$\degree$ for roll, pitch and yaw angle. The estimation errors for linear/angular velocity scaling factors are $6.0\times10^{-4}$ and $3.1\times10^{-3}$ separately. For scheme TC-LE-C, the calibration accuracy is worse. The translation errors are 0.11 m for X-axis and 0.09 m for Y-axis. The rotation errors are 0.24$\degree$, 0.16$\degree$ and 4.26$\degree$ for roll, pitch and yaw angle. The estimation errors for linear/angular velocity scaling factors are $2.0\times10^{-3}$ and $1.1\times10^{-4}$ separately. Compared to the TC-LE-C scheme, the calibration accuracy of TC-LE-OL is improved, especially at the translation part. The translation errors are 0.06 m for X-axis and 0.05 m for Y-axis. Other estimated parameters are almost the same. The rotation errors are 0.26$\degree$, 0.11$\degree$ and 4.60$\degree$ for roll, pitch and yaw angle. The estimation errors for linear/angular velocity scaling factors are $2.0\times10^{-3}$ and $3.1\times10^{-3}$ separately. 

Apart from GNSS-IMU lever-arm, a further insight is given into the initial values and the uncertainty of the calibration. As is shown in Fig. \ref{fig:sim_boxplot}, we disturb the initial value with Gaussian noise and perform 100 Monte-Carlo simulations for statistics. 
In the random simulation results, it can be observed that the choice of initial values does have a noticeable influence on the convergence process. Nevertheless, the translation extrinsic parameters are consistently estimated within 0.2 m accuracy, with the exception of the z-axis which remains unobservable under the given configuration. Moreover, the standard deviation (std) of the estimates provides a certain level of indication regarding the convergence behavior, but it does not fully cover the actual error distribution. This suggests that in practical applications the reported uncertainty should be treated as a reference rather than a strict bound.

\begin{figure}
    \centering
    \includegraphics[width=0.6\linewidth]{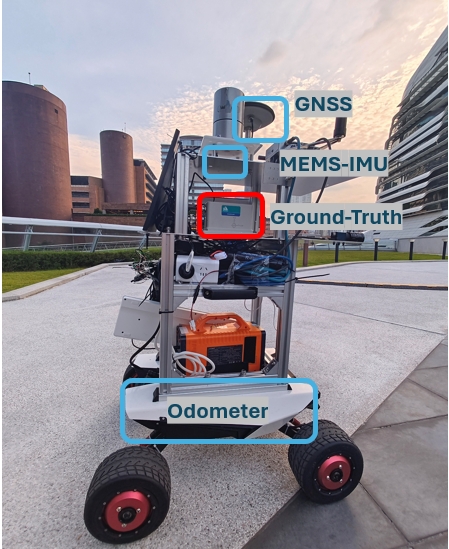}

        \caption{AGV platform for field data collection. Blue boxes denote the evaluating sensor and the Red box denotes the ground-truth system.}
    \label{fig:ugv}
\end{figure}

\begin{figure}
    \centering
    \includegraphics[width=1\linewidth]{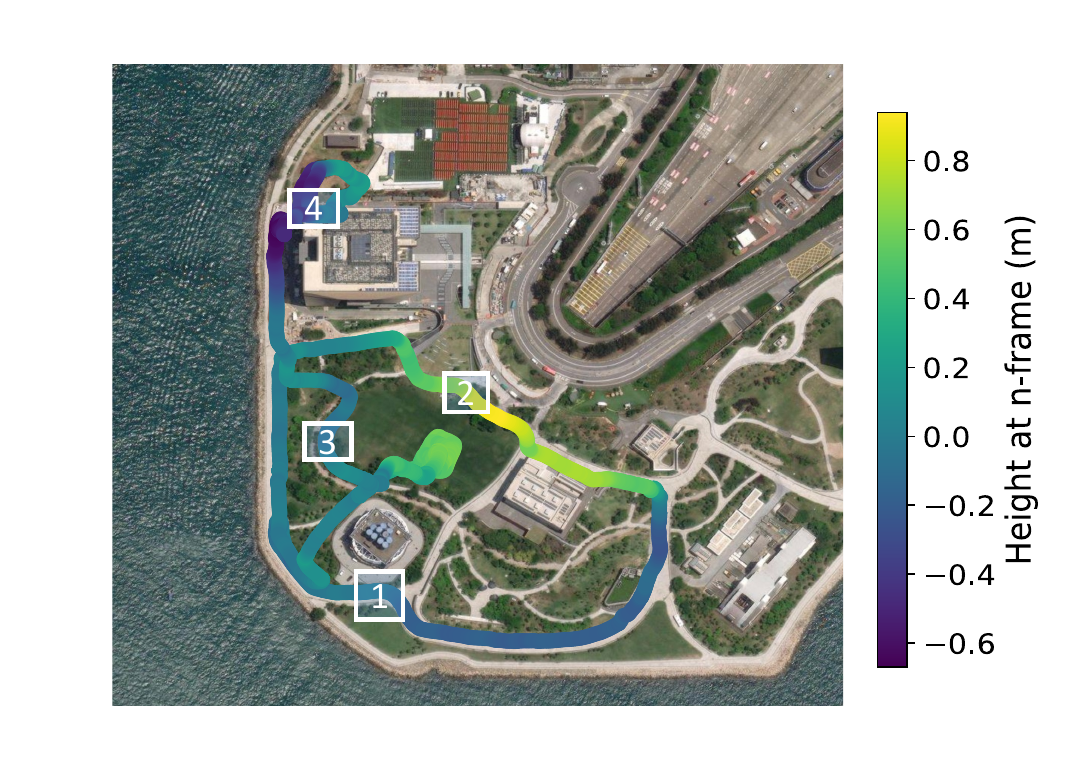}
    \caption{Google earth trajectory. The blue dots denote measurements over all the trajectory are used for calibration. The orange dots denote the trajectory where the measurements are used to test the performance of calibration results.}
    \label{fig:field_traj}
\end{figure}

\subsection{Discussion: Performance and Limitation of the Online Calibration}

The results show the observability of the proposed method. Most of the parameters are locally observable and converge in a several seconds under general planar motion and the precision of the calibration result is corresponding to the navigation state level. Notably, there is an error within 5$\degree$ for extrinsic yaw angle because of the low velocity and short IMU-odometer extrinsic translation, which might be decreased if we improve the precision of heading initial alignment. On the other hand, the extrinsic yaw angle is related to the estimated attitude state so high-level IMU can enhance the attitude estimation precision \cite{chen_estimate_2021}. For the translation along vehicle rotating direction, according to the result, the translation at the Z-axis of $v$-frame is not observable, proving that the lemma at Section \ref{sec:observability}. In fact, the lack of observability for height translation has little effect on the navigation performance because the odometer could sample in high-frequency and the motion in such a short time could be approximated as 2D motion. Moreover, to further lower down the impact of unobservable extrinsic parameters and raise the accuracy IMU-odometer integrated navigation, 2D motion detection is helpful and promising. We guess that if utilizing multiple IMU or perception sensors, the detection of 2D motion can be more accurate and the influence of IMU-odometer vertical translation parameter error can be further eliminated.

\subsection{Discussion: Impact of the GNSS-IMU lever-arm parameter disturbance}

As mentioned in the methodology, the proposed tightly-coupled factor graph can be divided into GNSS-IMU and IMU-odometer fusion parts. In the open-sky scenes, GNSS-IMU tightly coupled integration is able to perform positioning at centimeter level with predefined precise lever-arm parameters \cite{li_high-precision_2023}. Here, we analyze the impact of the GNSS-IMU lever-arm in the calibration process. In the proposed online calibration framework, the IMU-odometer extrinsic parameters are correlated with the navigation state. Thus the extrinsic transformation, especially the translation part will absorb the systematic error of the GNSS-IMU lever-arm. For the scheme TC-LE-C, it can be noticed that there are obvious offsets in X/Y translation, which is caused by GNSS-IMU lever-arm perturbation. Luckily, thanks to accurate GNSS measurements, GNSS-IMU lever-arm can also be estimated online in the same factor graph. For the scheme TC-LE-OL, the accuracy of estimated lever-arm comes to (0.01, 0.012, 0.12) m. Results in Section \ref{sec:sim_result} show that the final extrinsic transformation accuracy in TC-LE-OL is almost the same as the one in TC. Therefore, the GNSS-IMU lever-arm can be measured by tools at within centimeter level or estimated together in the estimator. Therefore, the impact of GNSS-IMU lever-arm with disturbance in centimeter level can be neglected after online calibration.

\begin{figure}
    \centering
    \includegraphics[width=0.75\linewidth]{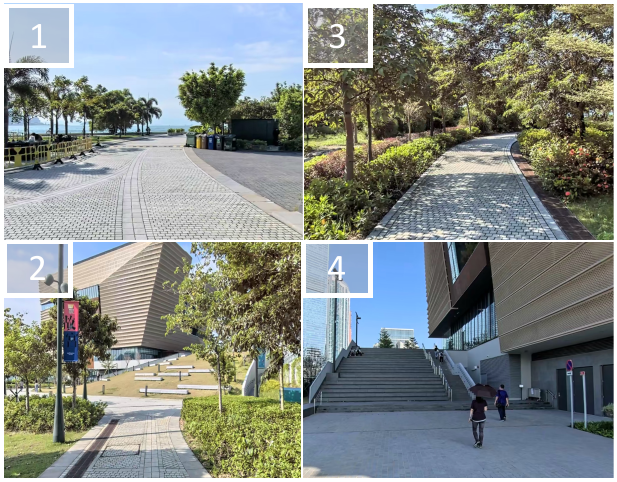}
    \caption{Four selected scenes for calibration evaluation.}
    \label{fig:field_selected_scenes}
\end{figure}

\begin{figure}
    \centering
    \includegraphics[width=1\linewidth]{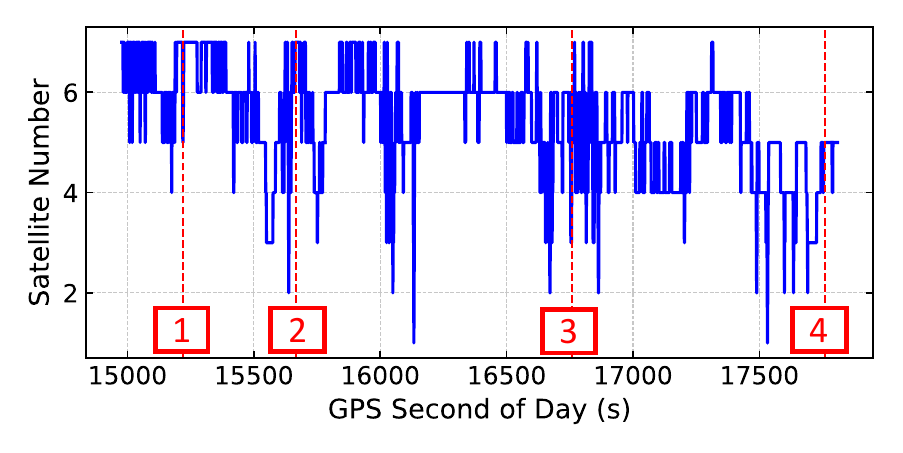}
    \caption{GPS Satellite number over time in the field test.}
    \label{fig:field_sat_pdop}
\end{figure}

\begin{figure}
    \centering
    \includegraphics[width=1\linewidth]{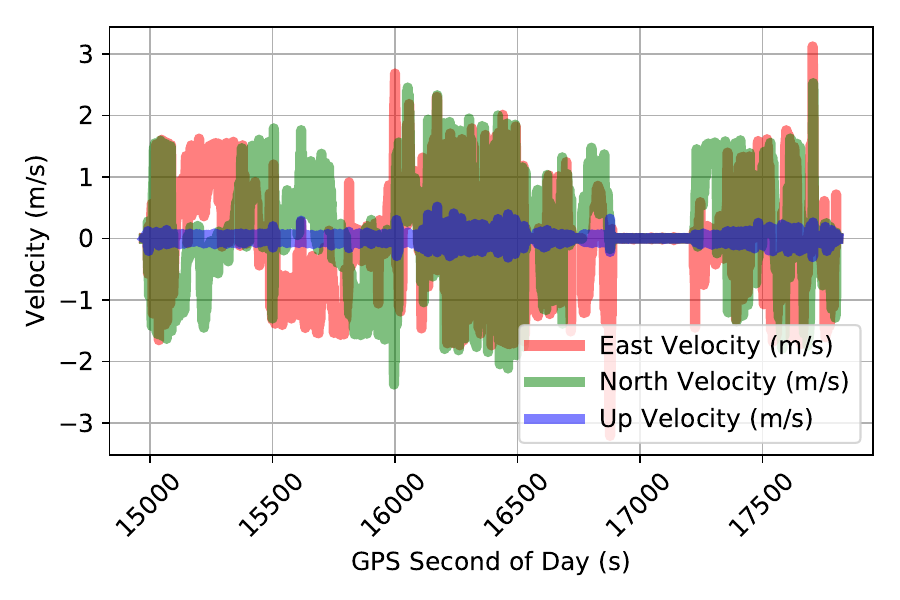}
    \caption{AGV velocity ground-truth at the local $n$-frame.}
    \label{fig:field_vel}
\end{figure}

\section{\label{sec:field}Field Test}
\subsection{Introduction of the field test}

\subsubsection{Data collection setups}
To verify the performance of the proposed calibration method, field experimental tests are conducted with a real-world ground robot dataset. To the best of our knowledge, Existing public ground robot datasets lack critical measurements from RTK base station \cite{yin_ground-fusion_2024} or odometer \cite{hsu_hong_2023}, making them unsuitable for this paper. Therefore, a self-assembled AGV robot is employed for real-world data collection on Nov. 4, 2024 in the Hong Kong West Kowloon Cultural park. As shown in Fig. \ref{fig:field_traj} and \ref{fig:field_selected_scenes}, the change of height is about 1.5 m. There are four typical scenes in the real-world tests, including open-sky, pathway covered by few trees, boulevard covered by more trees and half-sky blocking scenes. The satellite number of GPS is illustrated in Fig. \ref{fig:field_sat_pdop}, where the red lines denotes the epoch at four typical scenes. The employed ground robot is illustrated in Fig. \ref{fig:ugv}. During the experiment, GNSS measurements are collected by a low-cost u-blox-F9P dual-frequency receiver with a NovAtel GPS-702-GG antenna at 1 Hz; IMU measurements are collected by a MEMS-IMU Xsens MTi-30 at 400 Hz; odometer measurements are generated by an Agilex scout-mini mobile base with its open-source driver at 50 Hz. Besides, the ground-truth of navigation states is solved by NovAtel Inertial Explorer 8.9 software \cite{novatel_waypoint_2020} in the RTK/INS tightly coupled and smoothing filter mode using measurements from a NovAtel GNSS receiver and a high-precision tactical-level SPAN-CPT IMU. The key parameters of the MEMS-IMU and tactical IMU are presented in \cite{hsu_hong_2023}, which is also shown in TABLE \ref{tab:imu}. The baseline between the rover and the base GNSS station is about 4 kilometers, so the atmospheric delay could be eliminated clearly. 

In this work, we adopt the NHC when the mean angular velocity from the gyroscope is less than  $5\  \degree/s$  and the mean forward velocity from the odometer is greater than $1 \ m/s$ in one second. We also adopt the ZUPT when the mean angular velocity from gyroscope is less than $0.05 \ \degree/s$ in one second. The initial uncertainty of the IMU-odometer extrinsic parameters are 0.5 m and 5 $\degree$. The GNSS measurement uncertainty is $0.3 \ m$, $0.003 \ m$ and $0.3\ m/s$ for raw pseudo-range, carrier-phase and Doppler. The IMU bias uncertainty is set as $0.01 \ rad*\sqrt(Hz)$ and $0.02\ m/s*\sqrt(Hz)$. For the input of LC, we adopt the RTK result from RTKLIB with default configuration, including dual-frequency measurements, elevation weighting model and continuous ambiguity resolution mode. 
In this work, we adopt the NHC when the mean angular velocity from the gyroscope is less than  $5\  \degree/s$  and the mean forward velocity from the odometer is greater than $1 \ m/s$ in one second. We also adopt the ZUPT when the mean angular velocity from gyroscope is less than $0.05 \ \degree/s$ in one second. The initial uncertainty of the IMU-odometer extrinsic parameters are 0.5 m and 5 $\degree$. The GNSS measurement uncertainty is $0.3 \ m$, $0.003 \ m$ and $0.3\ m/s$ for raw pseudo-range, carrier-phase and Doppler. The IMU bias uncertainty is set as $0.01 \ rad*\sqrt(Hz)$ and $0.02\ m/s*\sqrt(Hz)$. For the input of LC, we adopt the RTK result from RTKLIB with default configuration, including dual-frequency measurements, elevation weighting model and continuous ambiguity resolution mode. 
All the data is collected and time-synchronized using a robot operation system (ROS) \cite{stanford_artificial_intelligence_laboratory_et_al_robotic_2018}, and the local computer time is aligned with the GPS time through Pulse Per Second (PPS) signal and the GPRMC message \cite{cao_gvins_2021}.
Note that the effect of time inconsistency is related to the motion of the vehicle. For example, for the ground robot used in this paper with a average forward velocity of about 1.5 m/s, a time misalignment of 10 ms only leads to 0.015 m, while for a car with a velocity of 20 m/s, the same time error could lead to 0.2 m. Therefore, it is necessary to consider the time misalignment instability. An similar work with a ROS software synchronization method has limited the time misalignment error below 30 ms \cite{hu_soft_2018}, equal to 0.045 m positioning error for our AGV. Therefore, the time misalignment is insignificant compared to our translation calibration error at decimeter-level. In our experiment, we adopt a sliding window size as 3 and the mean computation time is less than 0.5 s per epoch in an intel NUC 11 computer with i5-1135G7 CPU and 16 GB RAM . Therefore, it is suitable for real-time applications.

\begin{figure*}
    \centering
    \includegraphics[width=1\linewidth]{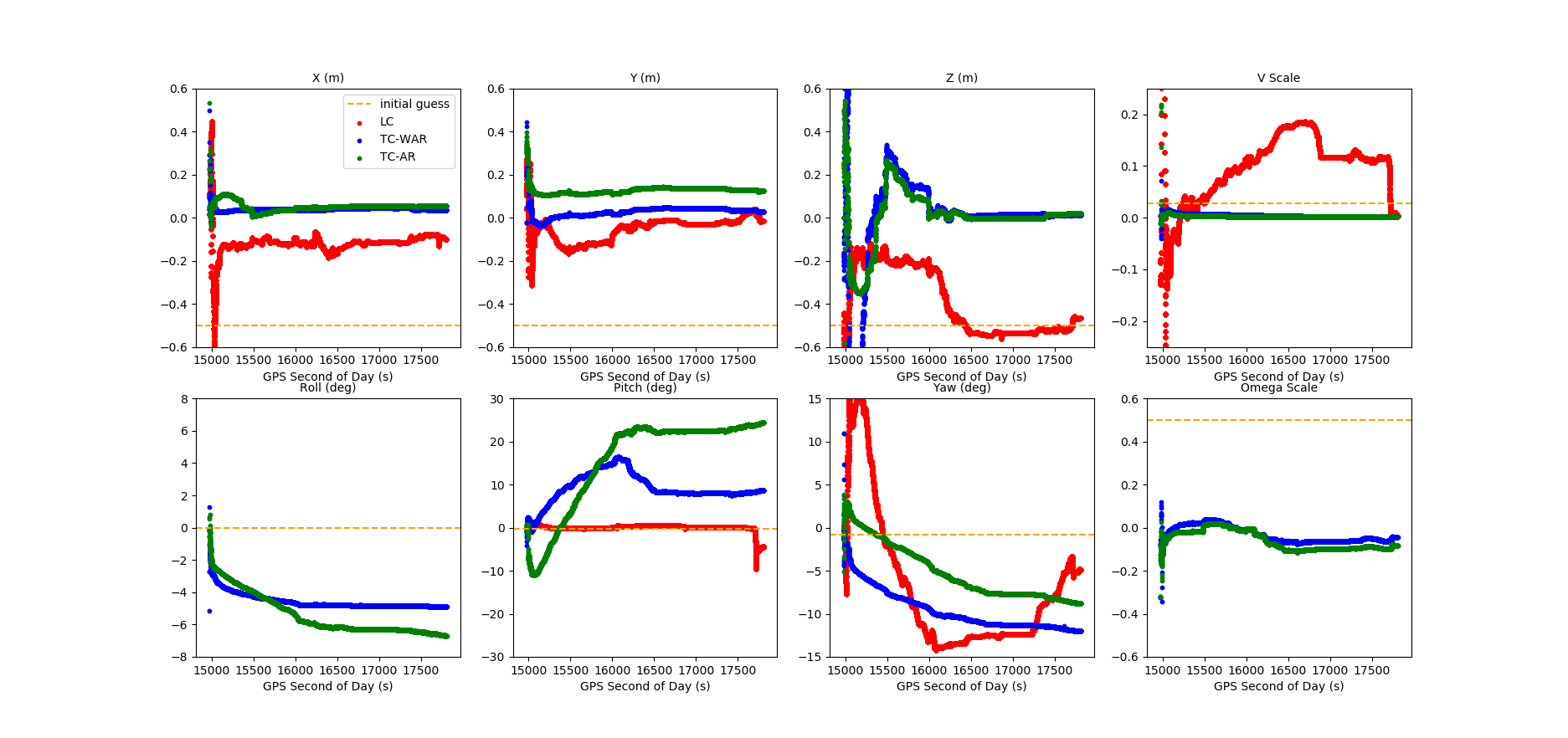}
    \caption{Calibration results of the field test, where orange dashed line denotes the initial guess; the red line denotes the loosely coupled (LC) method; the blue line denotes the tightly coupled method without ambiguity resolution (TC-WAR); the green line denotes the tightly coupled method with ambiguity resolution (TC-AR). Note that LC cannot estimate roll angle and omega scale since it only employs 1D odometer measurement (i.e., the linear velocity)}    \label{fig:field_calibration_result}
\end{figure*}

% \usepackage{booktabs}
% \usepackage{multirow}
\begin{table*}[]
\centering
\caption{IMU hardware configurations in real tests}
\label{tab:imu}
\begin{tabular}{@{}cccccc@{}}
\toprule
\multirow{2}{*}{IMU} & \multirow{2}{*}{Grade} & \multicolumn{2}{c}{Bias stability} & \multicolumn{2}{c}{Random walk}                  \\ \cmidrule(l){3-6} 
                     &                        & Gyro. ($\degree/h$  & Acc. (mGal)  & Ang. ($\degree/\sqrt(h)$ & Vel. ($m/s/\sqrt(h)$) \\ \midrule
NovAtel SPAN-CPT     & Tactical               & 1                   & 0.75         & 0.0667                   & -                     \\
Xsens-MTi30          & MEMS                   & 18                  & 0.015        & 1.8                      & 5.88                  \\ \bottomrule
\end{tabular}
\end{table*}
\subsubsection{Evaluation Scenes and Methods}

The velocity in $n$-frame is shown in Fig. \ref{fig:field_vel}. 
There are sufficient acceleration and turning to stimulate the extrinsic calibration process. .
To evaluate the robustness of our method under limited satellite observations, we only employ GPS to aid the IMU-odometer calibration.
The average satellite number is 5.39. To test the accuracy and robustness the proposed method, we employ a two-stage test to evaluate the performance of the proposed calibration method: (1) the calibration stage: parameters are estimated with measurements; (2) the testing stage: calibrated parameters at four typical scenes of the calibration stage are extracted and IMU-odometer positioning with simulated GNSS outage in about 1000 s is used to evaluate the calibration results. As is shown in Fig. \ref{fig:field_test_traj}, the test trajectory is a loop which is convenient for dead-reckoning based positioning evaluation. Based on these experimental settings, the compared methods are listed below:

(a) initial guess: coarse initial IMU-odometer parameters manually provided.

(b)	LC: the loosely coupled GNSS positioning constrained INS-odometer calibration \cite{bai_enhanced_2023} using RTK positioning from RTKLIB \cite{takasu_development_2009}.

(c)	TC-WAR: the proposed TC method without ambiguity resolution (WAR).

(d)	TC-AR: the proposed TC method with ambiguity resolution (AR).

\subsection{Estimation of IMU-odometer parameters at the calibration stage}

Fig. \ref{fig:field_calibration_result} shows the results of IMU-odometer extrinsic calibration at the calibration stage. 
% Since the online calibration is tightly related to localization , we also show the localization error of LC, TC-WAR and TC-AR in Fig. XX. With the motion . 
Since the LC method only utilize 1D odometer measurements, the roll angle and bearing rotating rate scale $s_\omega$ are not estimated. 
Notably, the fluctuation of the translation estimation is more obvious for the LC method. We consider the following reasons. On the one hand, the RTK positioning for the input of the LC method utilizing GPS requires at least three satellites and could degrade or fail with poor satellite geometry. On the other hand, the LC method can only deal with GNSS outliers at result level. Instead, the proposed TC method can process raw GNSS measurements with less information loss and therefore is more robust to GNSS outliers.
Despite the absence of the ground truth of the IMU-odometer parameters, we can still evaluate the calibration results indirectly using IMU-odometer dead-reckoning test. Specially, the pitch angle of both TC-WAR and TC-AR are drifting severely. We speculate that the pitch angle is highly related to the Z-translation during forward moving. Since the Z-translation error is not limited, the pitch angle of the IMU-odometer also absorbs that error. 
Moreover, since the GNSS-based sensor integrated navigation could be affected by environments, the calibration results at the four typical scenes mentioned above are evaluated.

\begin{table*}[]
\centering
\caption{IMU-odometer dead-reckoning positioning errors using the calibrated results from the calibration stage, including maximum absolute error (MAX) and root mean squares error (RMSE). The 'initial guess' column is the same across all four scenarios.}
\label{tab:field_pos_error}
\begin{tabular}{@{}p{2cm}p{2cm}p{2cm}p{2cm}p{2cm}@{}}
\hline
scene   MAX/RMSE (m)  & initial guess & LC & TC-WAR & TC-AR \\ \hline
1 (open)       & \multirow[c]{4}{*}{168.65/84.54} & 63.32/30.81  & 38.67/18.44 & \textbf{19.05/7.74}  \\
2 (few trees)  &                                  & 61.51/30.99 & 32.92/15.22 & \textbf{17.75/7.02}  \\
3 (more trees) &                                  & 51.40/31.53 & 34.27/15.80 & \textbf{25.30/10.81} \\
4 (half-sky)   &                                  & 30.48/14.17 & 36.66/17.12 & \textbf{27.08/11.77} \\ \hline
\end{tabular}
\end{table*}

\subsection{Performance of IMU-odometer localization with calibrated results at the test stage}
\begin{figure}
    \centering
    \includegraphics[width=1\linewidth]{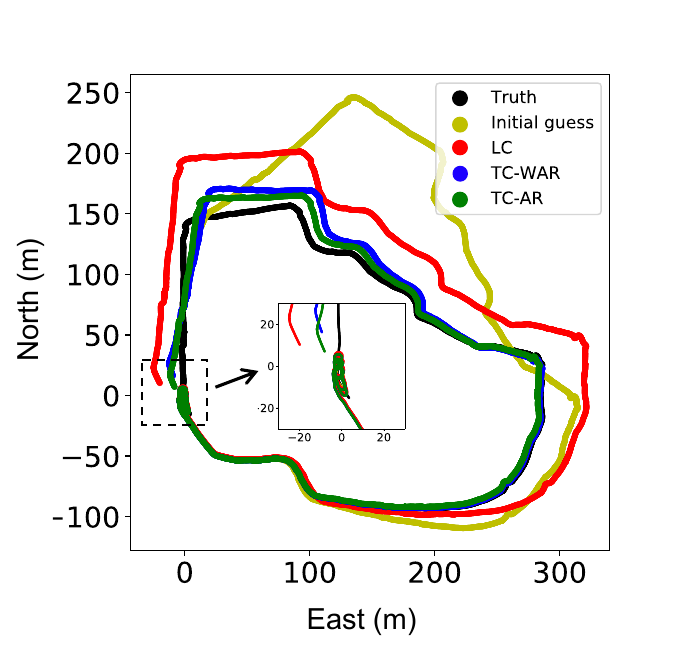}
    \caption{Horizontal trajectory comparison of IMU-odometer dead-reckoning test in scene 3 (more trees).}
    \label{fig:field_test_traj}
\end{figure}

In this section, the calibrated results are evaluated in a IMU-odometer dead-reckoning localization system. For fair comparison, we simulate GNSS outage after 100 seconds for a consistent initialization using GNSS measurements. Since the vertical translation at $v$-frame of the IMU-odometer extrinsic parameters is unidentifiable and may include large errors. We only compare the horizontal positioning errors in the following. It is worth noting that due to the inherent unobservability of the vertical extrinsic parameter and the coupling with extrinsic pitch and lever arm parameters, the TC method accumulates larger errors in elevation compared to LC. Nevertheless in the horizontal plane which is the practically relevant dimension the TC method consistently outperforms LC. For direct illustration, we plot the horizontal trajectory in $n$-frame in Fig. \ref{fig:field_test_traj} using the calibrated at scene 3 (more trees). The trajectories show that our proposed TC-AR method achieves the most accurate dead-reckoning performance in horizontal east-north directions.

Their horizontal positioning errors are presented in Table \ref{tab:field_pos_error}. Notably, MAX presents maximum absolute positioning error and RMSE denotes root mean square positioning error. Since the LC method does not output roll angle and the odometer scaling factor of angular velocity, we replace them with the same parameters from the initial guess during testing stage. The results show that MAX is 168.65 m and RMSE is 84.54 m for the initial guess scheme, which are the maximum ones among all compared schemes. Compared to that, the LC scheme gets smaller errors. From scene 1 to scene 4, the MAX shrinks from 63.32 m to 30.48 m and the RMSE reduces from 30.81 m to 14.17 m. For TC-WAR, the calibration is more accurate and robust. The MAX and RMSE of TC-WAR are 38.67 m and 18.44 m separately at scene 1. After that, the errors are reduced to 32.92 m and 15.22 m at scene 2 with longer stimulation. Nevertheless, when facing more challenging environments at scene 3 and 4, the MAX increases to 34.27 m and 36.66 m. Finally, the TC-AR provides the most accurate calibration results for this localization test. The trend of TC-AR follows a similar pattern as TC-WAR. TC-AR also achieves the best calibration at scene 2 with MAX of 17.75 m and the worst one at scene 4 with MAX of 27.08 m. 

Therefore, both TC-WAR and TC-AR are more accurate and robust than the initial guess and LC method, with TC-AR performing the best under these test conditions. TC-AR achieves up to 89.09\% and 71.14\% improvement in maximum horizontal positioning errors, respectively, compared to the initial guess and LC method.

\subsection{Discussion: Impact of the GNSS ambiguity resolution on the calibration performance}
In open-sky environments, only a few outliers in pseudo-range measurements are caused by multipath and NLOS effects. Then the GNSS can provide accurate relative ranging reference for calibration. In \cite{zhou_online_2022}, ionosphere-free (IF) GNSS measurements are employed to aid visual-inertial extrinsic calibration. Ambiguity resolution is not used in that work, likely because the authors utilized geodetic-grade receivers to refine the GNSS measurements. With such clean pseudo-range and carrier-phase data, the ambiguity variable is stable and the time-differenced carrier-phase is equal to a high-precision relative range measurement. Hence the extrinsic parameters can be estimated accurately, similar in principle to hand-eye calibration using accurate relative transform measurements \cite{giamou_certifiably_2020}.

In contrast, this paper focuses on low-cost sensors, which are more susceptible to measurement outliers. Although the double-differenced (DD) model eliminates most systematic errors, gross errors caused by the surrounding environment become more prominent in relative terms. Furthermore, low-cost GNSS receivers often adopt loose signal quality thresholds to maintain continuous tracking, allowing low-quality measurements to pass. To mitigate these environmental outliers, the OM method applied prior to AR removes a portion of the pseudo-range outliers, leading to an improvement up to 50.74\% in the maximum horizontal error when comparing TC-AR to TC-WAR. Nevertheless, cycle-slips or remaining gross errors may still lead to incorrect ambiguity resolution results, which could be researched in the future.

\renewcommand\stretch{1.5}

\section{\label{sec:conclusion}Conclusion}

In this paper, we presented an online IMU–odometer parameter calibration method for AGV localization using raw GNSS measurements. The proposed method is built on a tightly coupled factor graph optimization framework, which incorporates pseudo-range, carrier-phase, and Doppler measurements together with outlier mitigation and ambiguity resolution. We further conducted observability analysis, showing that two horizontal translation and three rotation parameters are identifiable, while the vertical translation remains unobservable. Both simulation and field experiments verified that our method significantly improves calibration accuracy and robustness compared with state-of-the-art loosely coupled approaches. Moreover, results indicate that the method can remain effective even without ambiguity resolution, enhancing its availability in practice.

Finally, some limitations still exist. The calibration problem is inherently non-convex, requiring reliable initialization for convergence. Future work will explore convex relaxation or certifiable optimization techniques to achieve global guarantees. 

\section*{acknowledgement}
The data processing of LC method \cite{bai_enhanced_2023} is supported by Dr. Shiyu Bai. Besides, the authors would like to thank the open-source software GICI-LIB \cite{chi_gici-lib_2023} from Shanghai Jiao Tong University. We develop our method based on this library.

\vfill
\end{document}